\documentclass{sigchi-ext}
\usepackage[T1]{fontenc}
\usepackage{textcomp}
\usepackage[scaled=.92]{helvet} 
\usepackage{graphicx} 
\usepackage{balance}  
\usepackage{booktabs} 
\usepackage{ccicons}  
\usepackage{ragged2e} 

\def\plaintitle{Why to Explain AI? Learning from Explanations: a Research Agenda} 
\def\emptyauthor{}
\def\plainkeywords{Explainable AI; AI complexity; Explanations \& Learning}

\title{Using Learning Theories to Evolve Human-Centered XAI: Future Perspectives and Challenges}

\numberofauthors{6}
\author{%
  \alignauthor{%
    \textbf{Karina Cortiñas-Lorenzo}\\
    \affaddr{Trinity College Dublin} \\
    \affaddr{Dublin 2, Dublin, Ireland} \\
    \email{cortinak@tcd.ie} } \alignauthor{%
    \textbf{Gavin Doherty}\\
    \affaddr{Trinity College Dublin}\\
    \affaddr{Dublin 2, Dublin, Ireland}\\
    \email{Gavin.Doherty@tcd.ie} }
    }

\definecolor{linkColor}{RGB}{6,125,233}
\hypersetup{%
  pdftitle={\plaintitle},
  pdfauthor={\emptyauthor},
  pdfkeywords={\plainkeywords},
  bookmarksnumbered,
  pdfstartview={FitH},
  colorlinks,
  citecolor=black,
  filecolor=black,
  linkcolor=black,
  urlcolor=linkColor,
  breaklinks=true,
}

\begin{document}

\CopyrightYear{2023}
\setcopyright{rightsretained}
\conferenceinfo{CHI'23,}{April  23--28, 2023, Hamburg, Germany}
\isbn{978-1-4503-6819-3/20/04}
\doi{https://doi.org/10.1145/3334480.XXXXXXX}
\copyrightinfo{\acmcopyright}

\maketitle

\RaggedRight{} 

\begin{abstract}
  As Artificial Intelligence (AI) systems continue to grow in size and complexity, so does the difficulty of the quest for AI transparency. In a world of large models and complex AI systems, why do we explain AI and what should we explain? While explanations serve multiple functions, in the face of complexity humans have used and continue to use explanations to foster learning. In this position paper, we discuss how learning theories can be infused in the XAI lifecycle, as well as the key opportunities and challenges when adopting a learner-centered approach to assess, design and evaluate AI explanations. Building on past work, we argue that a learner-centered approach to Explainable AI (XAI) can enhance human agency and ease XAI risks mitigation, helping evolve the practice of human-centered XAI.
\end{abstract}




\section{Introduction}
AI model complexity is soaring. Since 2018, it's estimated that the number of parameters in language models has increased by five orders of magnitude ~\cite{villalobos2022machine}. Powered by the release of models such as GPT-3 and coupled with growing computing power and vast volumes of data, the trend towards bigger models is only expected to accelerate ~\cite{large:scale:models}. Because training large-scale models is resource intensive, models are usually served pre-trained and fine-tuned for specific tasks, limiting the feasibility of tracing model predictions back to input training data. Additionally, AI models don't operate in a vacuum. In real-world settings, AI systems may comprise hundreds of models, data pipelines and integrations with several software components dynamically interacting with each other. In the face of such complexity, how can we compute faithful, complete and understandable AI explanations? Traditionally focused on explaining fully a given algorithm, many argue that Explainable AI (XAI) will soon become intractable ~\cite{sarkar2022explainable}. But should we aim at computing explanations that are faithful and complete? Why should we explain in the first place? Who is the explanation for? And what to explain? As a simile, we can consider human explanations. Irrespective of the complexity of the \emph{explanandum} or whatever is intended to be explained, human explanations have been used and continue to be used in our pursuit of understanding the world and controlling our environment ~\cite{keil2000explanation, lombrozo2006structure}. While explanations can serve multiple functions, both philosophers and psychologists have emphasized that explanations scaffold the kind of learning that supports adaptive behavior ~\cite{lombrozo2012explanation}, having a deep impact on cognition ~\cite{lombrozo2006structure}.   

While past research suggests that the main function of human explanations is to foster learning, XAI goals in the literature are varied, ranging from trustworthiness to regulatory compliance ~\cite{adadi2018peeking, das2020opportunities, arrieta2020explainable, saeed2023explainable}. Several works highlight the value of AI explanations to support knowledge discovery ~\cite{adadi2018peeking, saeed2023explainable}, learning about a domain ~\cite{liao2021human}, inference of causal relationships ~\cite{arrieta2020explainable} and model reconciliation ~\cite{sreedharan2021foundations}, but references to learning goals are scarce. Provided explanations are a transfer of knowledge ~\cite{miller2019explanation}, as proposed by ~\cite{kawakami2022towards}, engaging with AI explanations can be recast as a learning activity. Repurposing AI explanations as learning artefacts can have multiple benefits, including support of human agency and XAI risk mitigations, as well as enhanced actionability. In this paper, we review what we know about human learning and explanations and discuss how these learning theories might help inform the design of learner-centric AI explanations, as well as the implications of such as refocus.

\section{Explanations and Learning}
\subsection{How do humans learn?} 
Learning isn't the result of a single mechanism, but rather, it's understood to be influenced by a combination of several cognitive processes ~\cite{nokes2005comparing}. Given the diversity of mechanisms underlying learning, several theories abound focusing on various aspects of how learning takes place. These theories explain different ways by which people acquire knowledge and play a pivotal role in the design and implementation of education programs, influencing how teachers teach and learn to teach ~\cite{schunk2012learning}. Similarly, they can also influence how explanations can best support learning in different contexts:
\begin{itemize}\compresslist%
\item \emph{Behavioral theories:} learning is understood as the adaptation of behavior ~\cite{thorndike2017animal, skinner1954science}. Learner behavior is strengthen via reinforcers or weaken via punishers.
\item \emph{Cognitivism:} the focus is on the transmission of knowledge and how information is received, organised, stored and retrieved by the mind ~\cite{piaget1952origins, ausubel1978educational}.
\item \emph{Constructivism:} learners don't just receive knowledge from educators, but actively construct new knowledge and make meaning based on their experiences ~\cite{bada2015constructivism}.
\item \emph{Experiential learning:} learning is facilitated via interaction with the authentic environment where the learned skills are expected to be used ~\cite{kolb2014experiential}.
\item \emph{Humanistic theories:} concerned with personal growth and full development of human's potential, human learning theory argues that learning is self-directed and educators are just facilitators of it ~\cite{taylor2013adult}.
\item \emph{Reflective learning theories:} learning occurs when new knowledge becomes integrated into existing knowledge via reflection ~\cite{mezirow1990fostering, moon2013handbook}.
\item \emph{Social theories:} learning is situated within a particular domain of social practice, with social interactions and the community as the main facilitators ~\cite{bandura1977social, lave1991situated}.
\item \emph{Motivational theories:} motivation influences adult learning. For instance, the expectancy of success and rewards, as well as the learning goals and expectations matter ~\cite{weinerhuman, cross1981adults}.
\end{itemize}
Depending on the learning context, explanations can help support different learning theories. For instance, following a behavioral approach, explanations can be used as feedback to reinforce or discourage behavior, promoting learner's self-regulation ~\cite{afzaal2021explainable, rao2022explainability}. When adopting a reflective learning approach, explanations can trigger reflection rather than merely informing or prescribing. When adopting social theories of learning, they can be socially situated to facilitate reasoning ~\cite{ehsan2021expanding}.

\subsection{How do explanations influence learning?}
While learning involves different cognitive processes, explanations are thought to influence learning at three different stages ~\cite{lombrozo2012explanation}:

\begin{itemize}\compresslist%
\item \emph{Seeking explanations:} which explanations are sought constrain what is learned about the environment, with selectivity being driven by prior knowledge.
\item \emph{Receiving explanations:} the evaluation of explanations influences what is learned. The extent to which explanations provide a sense of understanding, not necessarily actual understanding, can impact the learning process.
\item \emph{Producing explanations (self-explanation effect):} past work has shown that generating explanations can be more effective in learning than actually receiving explanations ~\cite{williams2010role, chi1989self, rittle2006promoting}.
\end{itemize}


\section{XAI to support learning: future directions and challenges}
\subsection{\textbf{Using learning theories to assess XAI needs}}
\subsubsection{Opportunities}
Just as users often skip reading through privacy policies ~\cite{steinfeld2016agree}, past research has shown that people often ignore AI explanations ~\cite{naiseh2021explainable}. From a learning perspective, the search for explanations is driven by motivation and prior knowledge: useful explanations are expected to be epistemically valuable ~\cite{liquin2022motivated} and/or are associated with a given reward ~\cite{cross1981adults}. Motivational learning theories can offer a framework to structure the assessment of XAI needs, helping integrate into participatory methods aspects such as the attitude of the learner towards learning, the perceived likelihood and value of learning success, and factors such as learning incentives and time constraints. Because explanations are more likely to support learning if they are tailored to what the \emph{explainee} already knows ~\cite{lombrozo2012explanation}, constructivist and social learning theories can guide the identification of what is epistemically valuable for a given user in a given social-organizational context. Finally, taxonomies of knowledge such as Bloom’s ~\cite{bloom2020taxonomy} can help define learning objectives that can be elicited in later stages of evaluation.

\subsubsection{Challenges}
Learning via personal experience is a key tenet of theories such as constructivism and experiential learning. Assuming personal experience cannot be separated from learning, XAI needs might change over time and be highly dependent on a given individual, role, level of experience, etc. Additionally, incentives and time constraints might be highly dependent on context, requiring not only an initial assessment of requirements, but recurrent evaluations to achieve learning outcomes.

\subsection{\textbf{Delivering learning experiences via explanations}}
\subsubsection{Opportunities}
According to reflective learning theories, reflection matters and plays a pivotal role in adult learning ~\cite{mezirow1997transformative, moon2013reflection}. However, humans are also inclined towards simple explanations requiring low cognitive effort ~\cite{lombrozo2016explanatory}. Explanations that support learning don't necessarily lead to satisfaction ~\cite{buccinca2021trust}. In XAI, the drive towards simple and satisfactory explanations can lead to pitfalls such as unwarranted trust ~\cite{ehsan2021explainability}, over-reliance on AI outputs ~\cite{nourani2021anchoring} or automation complacency ~\cite{bertrand2022cognitive, schemmer2022influence}. Refocusing XAI on learning can help mitigate over-reliance risks. Because the goal is centered around learning outcomes, explanations should trigger further thinking, rather than be merely informative or prescriptive. Shifting towards a learner-centered approach requires refocusing our efforts on the \emph{explainee}, rather than the explanation, as well as a ruthless prioritization of those cases where the cognitive effort of engaging with explanations has high utility and can lead to the augmentation of human skills. From a learning angle, explaining AI system's algorithms may not even be necessary. Depending on the application, communication of model limitations, data informativeness or allowing users to triangulate system outputs (e.g. via probing ~\cite{hirsch2017designing, hook1998glass} or interacting with the system ~\cite{thieme2020interpretability}) could be sufficient to foster learning.

\subsubsection{Challenges}
Explanations can hinder learning if people are exposed to misleading regularities ~\cite{williams2013hazards}. Considering AI models might leverage spurious correlations in the data, how can the integration of factual knowledge help mitigate the risk of placebic explanations? Moreover, considering reductionist XAI approaches could lead to a sense of understanding rather than actual learning, how can we prompt users to reflect and question the information provided?

\subsection{\textbf{Evaluation of XAI methods using learning objectives and learning theory}}
\subsubsection{Opportunities}
When designing AI explanations to support learning, the evaluation of XAI methods can borrow from past research on education and cognitive sciences. For instance, from a behavioral learning perspective, XAI can be evaluated via application-grounded methods supporting behavior evaluation and from a cognitive learning approach, formative assessments can be used. Agreed learning objectives in the evaluation of XAI needs can also be resurfaced in this phase ~\cite{kawakami2022towards}.
\subsubsection{Challenges}
According to past research, learning is influenced by a combination of factors and contextual variables. Hence, due to the entangled nature of variables influencing learning, isolating the impact of explanations can be challenging. Additionally, application-grounded evaluation methods on the real task might not be directly accessible, requiring the use of proxy tasks and indirect learning measures.

\section{Conclusion}

Just as constructivist and humanistic learning theories facilitated a refocus of education from the teacher to the learner, so can a learner-centric approach help evolve the practice of HCXAI, helping refocus our efforts on the \emph{explainee} rather than the explanation. As we assess the value of explanations from a learner-centered perspective, we may realise that AI model explanations may not even be needed, moving from what's technically possible to what can really augment human skills and have a lasting impact on human learning and growth.

\balance{} 

\bibliographystyle{SIGCHI-Reference-Format}
\bibliography{sample}

\end{document}